# Textual Analysis for Studying Chinese Historical Documents and Literary Novels

[†]Chao-Lin Liu [‡]Guan-Tao Jin [↑]Hongsu Wang [§]Qing-Feng Liu
[↕]Wen-Huei Cheng [¦]Wei-Yun Chiu [¶]Richard Tzong-Han Tsai [ǂ]Yu-Chun Wang
[†]Department of Computer Science, National Chengchi University, Taiwan
[‡§↕¦]Department of Chinese Literature, National Chengchi University, Taiwan
[↑†]Institute for Quantitative Social Science, Harvard University, USA
[¶]Department of Computer Science and Information Engineering, National Central University, Taiwan
[ǂ]Department of Computer Science and Information Engineering, National Taiwan University, Taiwan
[†]Graduate Institute of Linguistics, National Chengchi University, Taiwan
[†]chaolin@nccu.edu.tw, [↑]hongsuwang@fas.harvard.edu, [¦]acwu0523@gmail.com,
[¶]thtsai@csie.ncu.edu.tw

## Abstract

We analyzed historical and literary documents in Chinese to gain insights into research issues, and overview[1] our studies which utilized four different sources of text materials in this paper. We investigated the history of concepts and transliterated words in China with the Database for the Study of Modern China Thought and Literature, which contains historical documents about China between 1830 and 1930. We also attempted to disambiguate names that were shared by multiple government officers who served between 618 and 1912 and were recorded in Chinese local gazetteers (地方志 /di4 fang1 zhi4/). To showcase the potentials and challenges of computer-assisted analysis of Chinese literatures, we explored some interesting yet non-trivial questions about two of the Four Great Classical Novels of China: (1) Which monsters attempted to consume the Buddhist monk Xuanzang in *the Journey to the West* (西遊記 /xi1 you2 ji4/, *JTTW*)*, which was published in the 16th century*, (2) Which was the most powerful monster in *JTTW*, and (3) Which major role smiled the most in *the Dream of the Red Chamber* (紅樓夢 /hong2 lou2 meng4/)*, which was published in the 18th century*. Similar approaches can be applied to the analysis and study of modern documents, such as the newspaper articles published about the 228 incident that occurred in 1947 in Taiwan.

## 1. INTRODUCTION

The immensely increasing availability of the digitized text material about Chinese history and literature offers great opportunities for researchers to take advantage of advances in computing technologies to conduct historical and literary studies more efficiently and at a larger scale than before, so *Digital Humanities* [7, 12] has emerged as a relatively new interdisciplinary field in recent decades. Researchers can employ techniques of information retrieval and textual analysis to extract and investigate information that are relevant to specific topics in their research. With the help of these computing technologies, researchers can obtain relevant information from a much larger data source than ever before, and this data collection phase can be completed a lot more efficiently as well.

---

[1] We report recent work and mention published results, in particular [3, 11, 17], to make this overview complete.

Software tools are useful not just for data collection. They can and should facilitate preliminary data analysis such that domain experts can spend their precious time and energy on more in-depth research, analyses, interpretation, and judgments.

Despite its relatively short presence in the research community, the ideas of conducting humanistic research with digital facilities have attracted the attentions and sometimes concerns of leading historians and philosophers in the worlds of western [6] and Chinese [16] languages.

In this paper, instead of discussing these developmental and philosophic aspects about digital humanities, we show how digital facilities can really support the studies of historical and literary documents in Chinese with four actual examples. Two of these research projects were conducted based on two different and large sources of historical text databases, and the other were based on two very famous classic Chinese novels.

The Database for the Study of Modern Chinese Thought and Literature (**DSMCTL**[2]) contains a wide variety of scanned documents and their text material about Chinese history and literature which were published between 1830 and 1930. With 120 million Chinese characters in the repository, DSMCTL has provided a crucial basis for the study about the history of concepts (觀念史 /guan1 nian4 shi3/[3]), and our research team has conducted a series of investigations about the establishment and variations of concepts, including "sovereignty" (主權 /zhu3 quan2/), "ism" (主義 /zhu3 yi4/), "Chinese People" (華人 /hua2 ren2/), and "Equality"(平等 /ping2 deng3/), with the help of software tools.

In many of our research projects, we relied on the temporal analysis of keywords for concepts and their co-occurrences. For example, to study the development of democratic concepts in China, we would search the Chinese translation of "democracy" in historical documents. In modern text, democracy is consistently translated to "民主" (/min2 zhu3/), so it is intriguing to look for "民主" for the study of democracy. However, the concept of democracy was a new concept to Chinese people, and people employed transliterated words to refer to democracy, i.e. "德模克拉西" (/de2 mo2 ke4 la1 xi1/), for some time. Hence, researchers would need to know this early embodiment of "democracy" in Chinese texts for their studies, and, to meet this need, we conducted a research for identifying transliterated words with a special book in DSMCTL.

*Difangzhi* (地方志 /di4 fang1 zhi4/) is a genre of official records published by local governments in China across many dynasties. Names and relevant information about government officers could be recorded in these local gazetteers. Extracting relevant information from *Difangzhi* and link the information about a particular person will help us strengthen the contents of the China Bibliographical Database Project (**CBDB**[4]) hosted by the Harvard University.

To this end, we need to tackle the problem of names that were shared by multiple persons. Some names are very popular than others. For instance, we have 29 records for 王臣 (/wang2 chen2/) and 29 records for 王佐(/wang2 zuo3/) in the our *Difangzhi* database. Few of them were owned by the same person, but most were not. Asking domain experts to compare and differentiate records for the same name in a collection of more than 110 thousand name records is quite beyond imagination because of time and costs. Hence, we employed computer programs to identify pairs of name records that might be owned by the same or different persons first to facilitate the name disambiguation task.

---

[2] http://dsmctl.nccu.edu.tw/

[3] Chinese words consist of one or more individual characters. For example, "人文" (/ren2 wen2/) is a Chinese translation of "humanities", and "人文" is a Chinese word that includes two Chinese characters. When we show a Chinese word the first time, we provide pronunciation information about its characters with Hanyu Pinyin followed by their tones in digits.

[4] http://isites.harvard.edu/icb/icb.do?keyword=k16229

In addition to analyzing historical documents, we explored the applicability of textual-analysis tools for Chinese literature. The most famous classic novels immediately came to our mind: the *Romance of the Three Kingdoms* (三國演義 /san1 guo2 yan3 yi4/), the *Journey to the West* (西遊記 /xi1 you2 ji4/), the *Water Margin* (水滸傳 /shui3 hu3 chuan4/), and the *Dream of the Red Chamber* (紅樓夢 /hong2 lou2 meng4/). All of them have been translated into English and other languages. Using these novels as the bases for our illustrative studies will be appreciated more easily by the domain experts and ordinary people.

In this paper, we report our work with the *Journey to the West* (***JTTW***, henceforth) and the *Dream of the Red Chamber* (***DRC***, henceforth). We chose to work on three questions whose answers were not immediately obvious for readers who read *JTTW* and *DRC* even not just once.

For *JTTW*, we would like to find out the monsters which attempted to consume the Buddhist monk Xuanzang, who is arguably the most important role in *JTTW*. In *JTTW*, many believed that consuming the monk will make one immortal, so a number of monsters chased after the monk for immortality. Also about the monsters in *JTTW* is which monster was the powerful. Instead of reading the stories to compare, we offer a qualitative but simple approach to respond to this interesting question.

For *DRC*, we wondered the answer to the question: who was the one that smiled most frequently among the three most important characters in the novel, i.e., 寶玉 (/bao3 yu4/), 黛玉(/dai4 yu4/), and 寶釵(/bao3 chai1/)?

We elaborate on each of these aforementioned studies in separate sections along with discussions about limitations of our current approaches, and wrap up this paper with concluding remarks and some future work.

## 2. THE DATABASE FOR THE STUDY OF MODERN CHINESE THOUGHT AND LITERATURE

The Database for the Study of Modern Chinese Thought and Literature (**DSMCTL**) contains more than 120 million Chinese characters. This relatively large database serves as a good resource for research, though it is quite formidable for anyone to read all of its contents.

Software tools offer two levels of assistance and prove to be instrumental for the efficiency and effectiveness in our work. We have built tools which help historians identify and extract potentially relevant text material for further in-depth research. We also implemented tools which allow historians to examine statistical properties of important keywords and their co-occurrences[5].

In a typical study, historians initiated a research problem and provided a list of relevant seed keywords for the target problem. Historical documents were then identified and extracted from DSMCTL based on these initial seed keywords. Given this initial set of extracted documents, historians could browse them and then selected the documents that were really relevant to the target problem.

We then employed computing tools to help us find very frequent words ("**pseudo words**, henceforth) in these selected documents, and the historians could inspect the contexts of these pseudo words to pick a set of new keywords from these pseudo words. If the historians were curious about the significance and relevance of these new keywords to the target problem, we could extract documents that contained these new keywords for the researchers to inspect. This iterative step of identifying important keywords and extracting relevant documents can continue many times as needed.

With the selected keywords, we could compute their statistical properties. Temporal analysis of the keyword frequency is the most fundamental tool. This analysis provides some visual trends about the appearance of a keyword over time. The ups and downs of keyword frequencies may suggest interesting historical events hidden in the text records, and often triggers new ideas for the study.

Figure 1 illustrates a temporal analysis for the keywords that are related to the movements of constitutional monarchy in China between 1905 and 1911. The curves were drawn based on the statistics collected from the official documents of the central government. The changing trends of the curves indicated the main activities of the central government.

[5] TaiwanDH (Taiwan Digital Humanities): https://sites.google.com/site/taiwandigitalhumanities/

In addition, we also ran temporal analysis of co-occurrences (commonly referred to as "collocations" in computational linguistics) of keywords. A collocation usually refers to a pair of words, i.e., bigrams, which appeared within a selected range of text, e.g., a sentence. Yet, there were no reasons which prevented us from analyzing trigrams and more complex contexts. The actual meaning or semantics of a word was influenced by its context [5], so the temporal analysis of collocations provided a better opportunity to discover more precise implications of the appearance and/or missing of keywords in the historical documen ts.

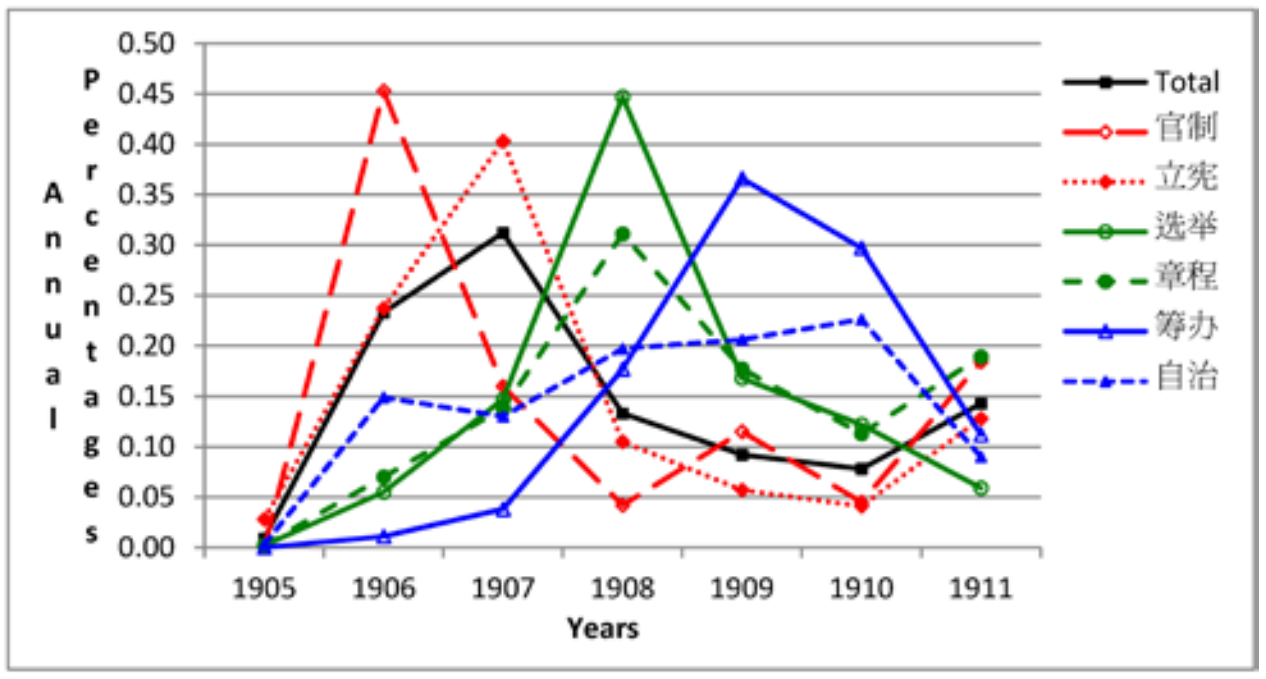

**Figure 1. A temporal analysis of keywords for the study on the movement of constitutional monarchy in late Qing dynasty [11]**

Figure 2 shows the changing trends of selected collocations of keywords for the study on the formation of "Chinese People". The peaks of the curves correspond to historical events that can be found and verified in Wikipedia.

An obvious barrier in conducting the analysis of collocations was that there were a humongous number of collocations to be examined. With 100 interesting keywords, for example, a historian might have to examine at most 10,000 collocations (bigrams). At this moment, we deployed software tools to help historians examine and records the original text of these collocations so that they could efficiently select the collocations that attracted their attention for further study.

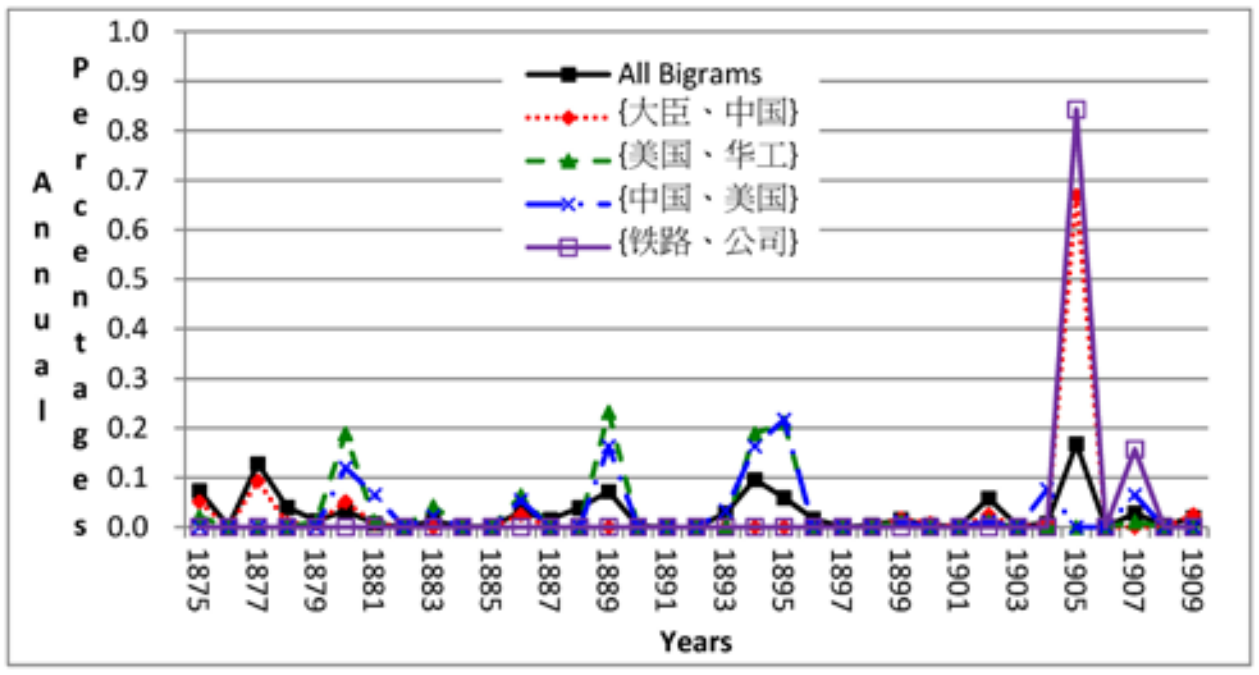

**Figure 2. A temporal analysis of collocations of keywords for the study on the concept formation of "Chinese People" [11]**

With these supportive software facilities, historians can explore the text material contained in DSMCTL with better efficiency and probe into a much larger amount of texts that were almost not possible before. After carefully identifying important keywords and collocations with the help of the statistical analyses, historians can focus on the reading and interpretation of text materials that were really related to the target problem.

Researchers participating in the DSMCTL project have employed these computing tools and procedures to investigate several historical issues. We studied the changing usage of "Sovereignty" (主權 /zhu3 quan2/) between 1860 and 1928, and looked into the migrating collocations of "ism" (主義 /zhu3 yi4/) between 1896 and 1928. We examined the historical documents to find the burgeoning concept about "Chinese Labor" (華工 /hua2 gong1/), "Chinese Businessman" (華商 /hua2 shang1/), and "Chinese People" (華人 /hua2 ren2/) from 1875 to 1909.

### *2.1. History of Concepts*

More specifically, experiences gained in linguistic research show that "You shall know a word by the company it keeps" [5]. By analyzing the changing collocations of "Equality" (平等/ping2 deng3/), we verified the evolution of the concept about "Equality" in the Chinese society in three periods: 1898-1900, 1901-1914, and 1915-1924, that was proposed and discussed in [3].

Tables 1 and 2 show the statistics of the frequencies of keywords that collocated with the word "Equality" in different periods. We can see that, in different periods, different sets of words collocated with "Equality" more often than others, and these different sets of collocations and their original contexts altogether implied different concepts of "Equality". At one stage, people sought equality of the nation, when the Qing dynasty was really weak and was invaded by the Western powers. At another stage, people were bothered by the inequality between the public and the private sectors. Equality among the ordinary people became an issue after the nation turned democratic.

### *2.2. Transliterated Words in Historical Documents*

We have developed techniques to identify transliterated words in Chinese historical documents [17]. Concepts represented by words like "president" and "democracy" were new to Chinese, and how people recorded these concepts in Chinese words are important for the study of these concepts in Chinese history. Evidence indicated that Chinese transliterations of these new concepts may vary over time, so it is important though difficult to find all variants for referring to the same concept in Chinese historical documents.

Table 1. Frequencies of frequent collocations of "Equality" (平等) for the period between 1898 and 1900 [3]

| | 1898-1900 | 1901-1914 | 1915-1924 |
|---|---|---|---|
| 西人 | 43 | 10 | 9 |
| 強權 | 39 | 14 | 12 |
| 萬國 | 28 | 21 | 5 |
| 生滅 | 23 | 2 | 0 |

Table 2. Frequencies of frequent collocations of "Equality" (平等) for the period between 1901 and 1914 [3]

| | 1898-1900 | 1901-1914 | 1915-1924 |
|---|---|---|---|
| 權力 | 7 | 121 | 25 |
| 道理 | 1 | 59 | 9 |
| 平等之地位 | 0 | 58 | 2 |
| 服從 | 6 | 53 | 13 |
| 立憲 | 0 | 100 | 12 |
| 君主 | 7 | 88 | 17 |
| 滿漢 | 0 | 59 | 0 |
| 漢人 | 2 | 54 | 2 |
| 滿洲 | 0 | 51 | 5 |
| 政體 | 5 | 50 | 7 |
| 同胞 | 8 | 56 | 12 |
| 眾生 | 9 | 51 | 7 |
| 強弱 | 5 | 52 | 8 |

We conducted our  study with a special book, 海國圖志 (/hai3 guo2 tu2 zhi4/, ***HGTZ*** henceforth) that contains many transliterated words, and the transliterations are already manually marked by domain experts in China. *HGTZ* was published in the Qing dynasty (ca. 1841AD), co nsists of 100 chapters, and contains about 680 thousand characters.

Since the transliterated words may not be recorded in any lexicon, we have to look for transliterations from raw strings. After obtaining strings that appear more than twice, we sifted the candidate strings with different filters. The goal was to reduce the number of candidate strings that will be manually checked by domain experts for transliterated words.

Like a traditional task of information retrieval, we would wish to achieve high precision and high recall rates for this process. Removing the candidate strings aggressively may save the domain experts a lot of time for manual filtering but may result in poor recall. Keeping a lot of candidate strings for manual inspection may boost the recall rate at the cost of poor precision rate, and that would also make the domain experts spend a lot time to complete the selection.

We have three different types of filters in the current work. The first one is remove strings that frequently appeared in non-historical documents, e.g., literatures such *the Dream of the Red Chamber*. It is quite unlikely that transliterated words would appear frequently in literary novels.

The second type of filter is to consider the special features of Chinese pronunciation and word formation patterns. The phoneme and lexical patterns of transliterated words may not differ very much from ordinary Chinese words because they will be used in ordinary Chinese texts.

The third type of filter is to consider the textual contexts of the transliterated words. Since the transliterated words in *HGTZ* were manually marked, we could extract higher-level linguistic features about the transliterated words and employ machine learning methods to mine the rules about the textual contexts in which transliterated words appeared, and then applied the rules to rank the candidate strings.

We ran experiments on a test set of more than 200 thousand candidate strings. Only 57,024 of them passed the first and the second type filters, while the recall rate was at 76.54%. We then ranked the remaining candidate strings with the machine-learning based method, and found that 96.14% of the leading 500 candidates were indeed transliterated words.

The performance of our filtering and ranking methods may look satisfactory from the perspective of computer science. However, it is possible for a historian to demand higher recall rates because unpredictable problems may ensue the omission of any transliterated words.

Transliterated words that appeared only once in the source text is another problem that we have not handled efficiently yet. In fact, there is one such instance in *HGTZ.* At this moment, a string must appear at least twice to be considered as a candidate transliteration. If we would consider strings that appear only once, the number of the candidate strings will increase dramatically and that will lead to big challenges to our data processing capacity.

## 3. DIFANGZHI (CHINESE LOCAL GAZETTEERS)

Currently, the China Biographical Database Project (CBDB) hosted by the Harvard University offers free download of a database for Chinese biographical information. Enhancing the contents of the CBDB database is an ongoing task, and a good source of additional information may come from the *Difangzhi*, which is a large collection of local gazetteers compiled by local governments across many dynasties in China.

To this end, we have employed the techniques of regular expressions to extract information about individuals, and, at the time of this writing, we obtained more than 110 thousand records for about 84000 different names [12]. Quite a few of these records are for the same names, e.g., we have 29 records for the name 王臣 (/wang2 chen2/). Some records for the same name may belong to the same person, and some do not.

Before we can augment the CBDB database with the information from *Difangzhi*, we will have to determine whether or not the owners of the *Difangzhi* records with the same name are the same or different. We call this task a *name disambiguation* task, and we are implementing an algorithm laid out by Bol [1]. This algorithm considers many factors in a name record, including birth place, entry into office (入仕方法 /ru4 shi4 fang1 fa3/), office posting (職官 /zhi2 guan1/), alternate names (字號 /zhi4 hao4/), service location (任職地點 /ren4 zhi2 di4 dian3/), service periods (任職時間 /ren4 zhi2 shi2 jian1/), etc., and we compare these factoids of two name records to compute a score for their similarity.

Temporal and spatial information are two important categories of information for the task of name disambiguation. The information about service periods in two records, for example, may help us differentiate two persons with the same name when the service periods were far apart.

Spatial information includes the birth places, the service locations, and the publication addresses of the *Difangzhi* books. If two name records have the same values for these items, they are more likely to belong to the same person.

The comparison of two location names is a challenging problem itself. When two location names do not match literally, they may have other indirect relationships. Two locations might belong to the same governing location name; e.g., 龍川縣 (/long2 chuan1 xian4/) and 惠川縣 (/hui4 chuan1 xian4/) were two different locations, but they were governed by 惠州府 (/hui4 zhou1 fu3/) in the Qing dynasty. Hence, they could be considered related. In addition, one of the two location names may be a governing location name of the other. For instance, 宜寧縣 (/yi2 ning2 xian4/) does not match 處州府 (/chu3 chou1 fu3/) literally, but 宜寧縣 was a subarea within 處州府 in the Qing dynasty. Moreover, two location names that are literally different may be considered as a match if their GPS coordinates are sufficiently close to each other. We obtain the belonging and GPS information about location names from the China Historical GIS[6] service hosted by the Harvard University.

Missing values form a major hurdle for the comparison between records. The contents of *Difangzhi* books did not adopt a consistent format, and it is very normal to find many of the factors in a name record not having values. This missing value problem posts a major challenge for the name disambiguation task.

After comparing the factoids in a pair of records for the same name, we can estimate the degree of match of the pair, and calculate a similarity score for the pair. For pairs that have high similarity scores, we may be able to assume that the pairs belong to the same person. Pairs that have very high dissimilar scores may belong to different persons. Pairs that have marginal similarity scores may require human inspection to determine their relationships.

We have implemented a good prototype system for comparing the pairs and for calculating the similarity of the pairs. We have identified nearly 40,000 pairs of records that have non-zero similarity scores. The results of this name disambiguation task still need to be judged and evaluated by domain experts, which we have not completed yet.

## 4. JOURNEY TO THE WEST

We have also demonstrated the applications of software tools to the study of Chinese literature by attempting to efficiently find answers to interesting and non-trivial questions about the contents of famous Chinese novels.

---

[6] http://www.fas.harvard.edu/~chgis/

*The Journey to the West* (西遊記 /xi1 you2 ji4/) consists of 100 chapters and has more than 713 thousand characters. The first question for *JTTW* is: Who were the monsters that wanted to consume the Buddhist monk Xuanzang (唐三藏 /tang2 san1 cang4/) in *JTTW*? The answer to this question was not immediately obvious even to the readers who may have read the novel multiple times, and the goal was to find the complete list of the monsters as soon as possible, while not reading *JTTW* completely. Except having completing the task in a short time, we actually discovered new answers that domain experts did not know of.

To solve this problem, we employed software tools to extract and identify statements and chapters that might be relevant to the eating of the Monk. The extraction of statements was based on some keywords selected by a novice who is a well-educated native speaker of Chinese but has never read *JTTW*. In order to eat the Monk, the monsters must capture the Monk and somehow made the Monk ready to be eaten by a process of cooking. Thus, there should be some specific words that would be used to describe this capturing, cooking, and eating process.

After extracting the candidate statements, the novice picked the statements that he believed to be related to the actions of capturing and eating the Monk. Then, he browsed the parts of the chapters that are close to the chosen statements to find the names of the monsters. Since the goal was just to find the names of the monsters, it was not necessary to read the whole stories in chapters.

While selecting the relevant statements and browsing the nearby paragraphs, the novice may find new keywords that he believed to be helpful for extracting more relevant statements. In such cases, the novice could use the software tools to find extract more candidate statements for further inspection. The search process is essentially iterative.

By using the software tools that facilitated the extraction of candidate statements and browsing of relevant contents in chapters, the novice did not have to read *JTTW* completely to spot the names of the monsters who attempted to eat the Monk.

Results of this experiment proved the potential of using digital tools to study literature. To compete for efficiency, the novice spent less than 10 hours to accomplish the task, though he could have inspected the text more carefully. *JTTW* has more than 713 thousand Chinese characters, and it was impossible to examine *JTTW* carefully in such a short time. One has to read and understand nearly 20 Chinese characters[7] a second to reach this speed.

The novice found 20 monsters and their names. The list of monsters thus identified contained two monsters that a well-known domain expert about *JTTW* did not know of, where the expert was the president of a research-oriented university in Taiwan. The ability to discover new answers that domain experts did not know before should be considered as a very attractive potential of using computational tools for textual analysis. On the other hand, the novice's list also missed one monster in the expert's answers.

Using the typical precision and recall metrics to measure the "outputs" of the novice, who worked with the help of the software, and the expert. The precision rates of the novice and the expert are both 100%. The recall rate of the novice is 21/22, and the recall rate of the expert is 20/22.

In reflection, the main reason that made the novice fail to find the missing monster was a missing keyword in the chosen keywords based on which the software extracted candidate statements for the novice to inspect. In this situation, a good starting point, i.e., the keyword list, materially influenced the search results, and should be prepared and supported by a more powerful mechanism.

The second question for JTTW is: Who is the most powerful monster in JTTW? When one first encounters this challenge, s/he would be shocked because answering this question almost requires one to read JTTW again.

This question is not difficult if we can consult some relevant knowledge. Figure 3 plots when some monsters appeared in the chapters of JTTW. Figure 4 depicts when some Dao (道教, /dao4 jiao4/) masters appeared in the chapters. The horizontal axes show chapter IDs with a prefix "d". Namely, "d001" represents chapter 1. Since normally the Dao masters appeared in the chapters to fight against the monsters, we could rank the powers of the monsters by the powers of the Dao masters. It may be hard to estimate the powers of the masters, but the powers and ranks of the Dao masters are known (or believed) to the human experts.

---

[7] 713000 / 10 hours/ 60 min per hour / 60 secs per min ≈ 19.8

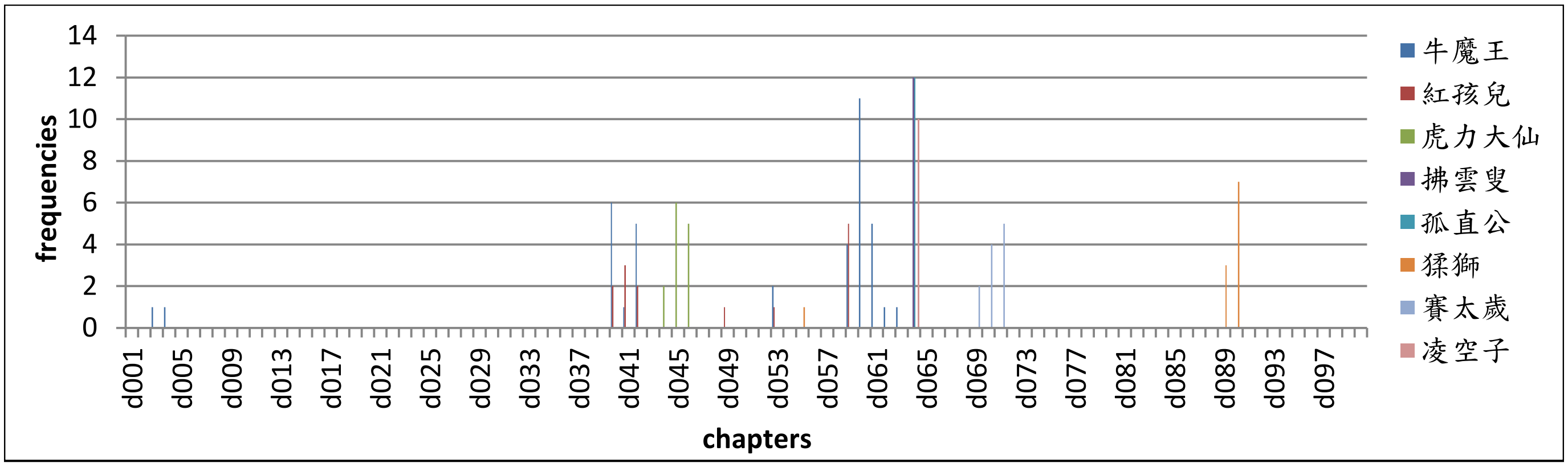

**Figure 3. Appearances of some monsters in *JTTW***

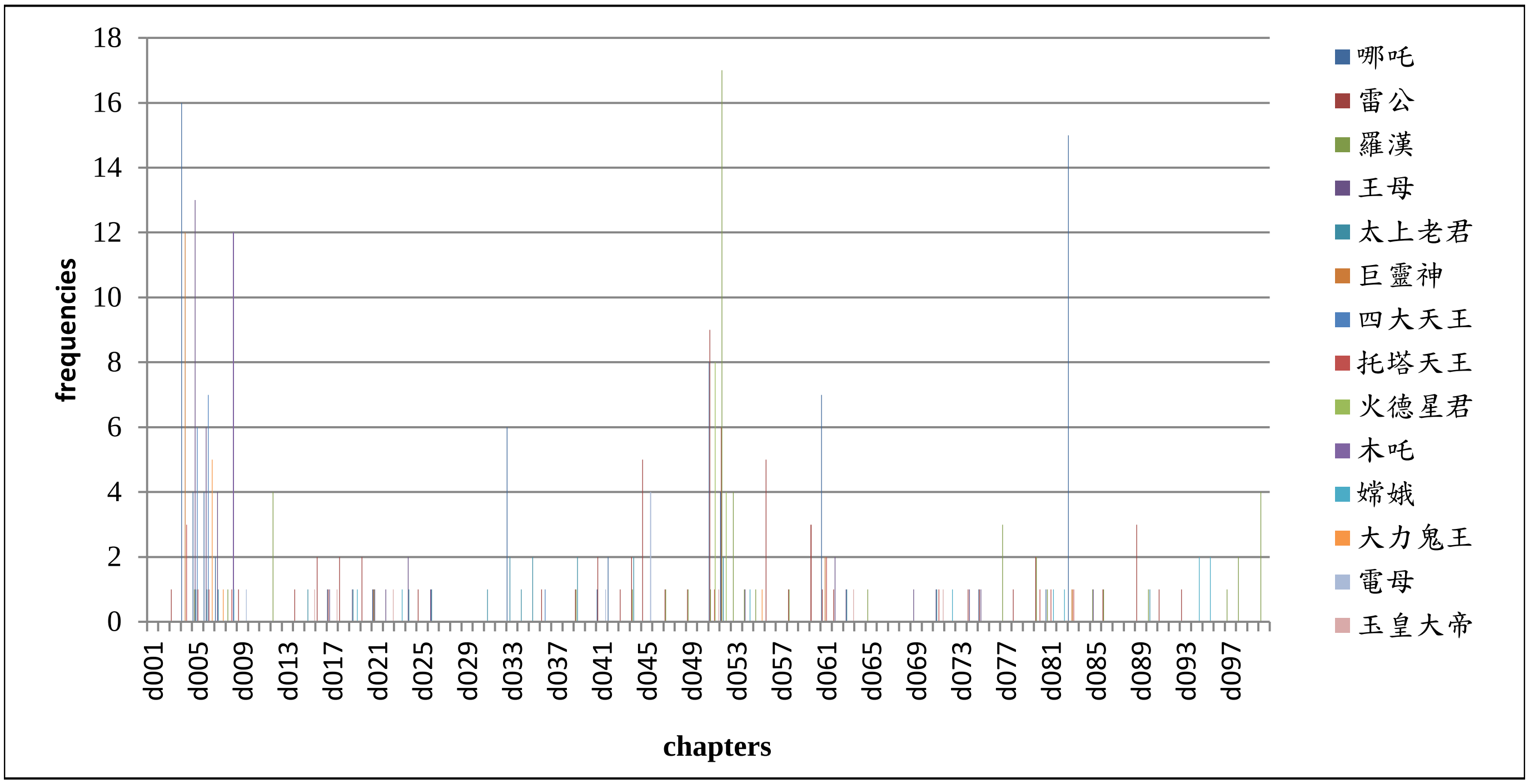

**Figure 4. Appearances of some Dao masters in *JTTW***

Hence, aligning the chapters of Figures 3 and 4 by the chapters would provide clues to the powers of the monsters, given the information about the powers of the Dao masters. More masters who are more powerful were needed to conquer the most powerful monster.

## 5. Dream of the Red Chamber

The most well-known research topic about *DRC* is about its authorship, cf. [15]. *DRC* has 120 chapters, and many believe that the first 80 chapters and the last 40 chapters were produced by different authors. Rather than working on this relatively popular research topic, we invented a new question about *DRC* – Who among 寶玉 (/bao3 yu4/, **BY**, henceforth), 黛玉(/dai4 yu4/, **DY**, henceforth), and 寶釵(/bao3 chai1/, **BC**, henceforth) liked to smile most in *DRC*?

Calculating the frequencies of "smiling" is a direct way to find out the answer, and Figure 5 shows the frequencies of BY, DY, and BC 笑道 (/xiao4 dao4/, smile and say) in each chapter. The chart illustrates an alternative form of temporal analysis that we had shown in Figure 1, if we would consider the progressive chapters as time stamps. The curves suggested that BY is the person who smiled more frequently before the 36th chapter. Furthermore, the chart also suggests that BY smiled more frequently before the 36th chapter than after the 36th chapter.

Directly computing the frequencies of smiling provides a simple answer, and it is easy to implement programs to calculate the raw frequenci es of any events of interest. Nevertheless, this simple solution may not provide a solution that satisfies everyone.

The reason is quite simple: A person is more likely to smile if s/he appears more frequently. Not surprisingly, there are short and long chapters in a realistic novel. In a longer chapter, it is more likely for some characters to appear and to smile, all else being equal. Hence, it is not really reliable to claim that a character smiles more often in certain chapters based only on the raw frequencies of smiling.

Indeed, a more basic question before we try to answer the original question is how we define “smile more often”. A conceivable alternative of the raw frequencies of smiling is to divide the raw frequency of smiling by the frequency of a character’s appearance. Figure 6 shows the proportions of BY’s, DY’s, and BC’s smiling in individual chapters given their appearances. One may receive the impression that BC, not BY, was the person who liked to smile most. Although the absolute frequency of BY’s smiling was higher, BY did not smile as often as BC when they appeared.

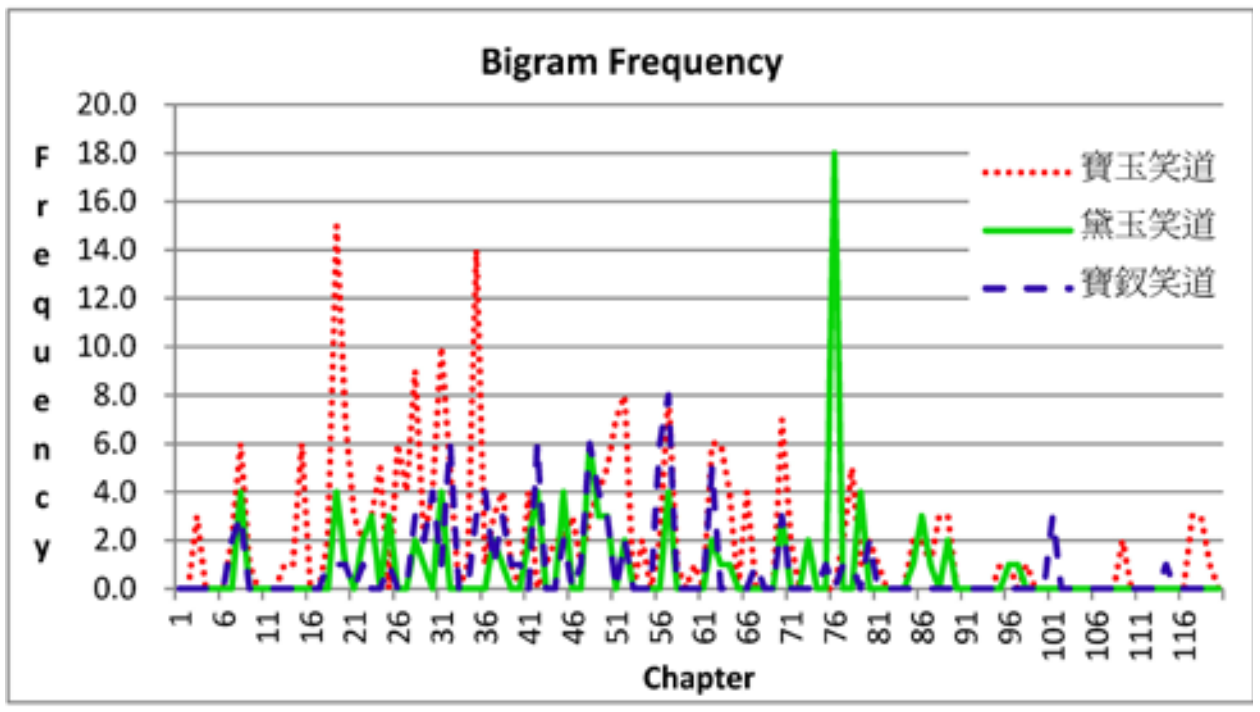

**Figure 5. Frequencies of smiling of three main characters in *DRC* [11]**

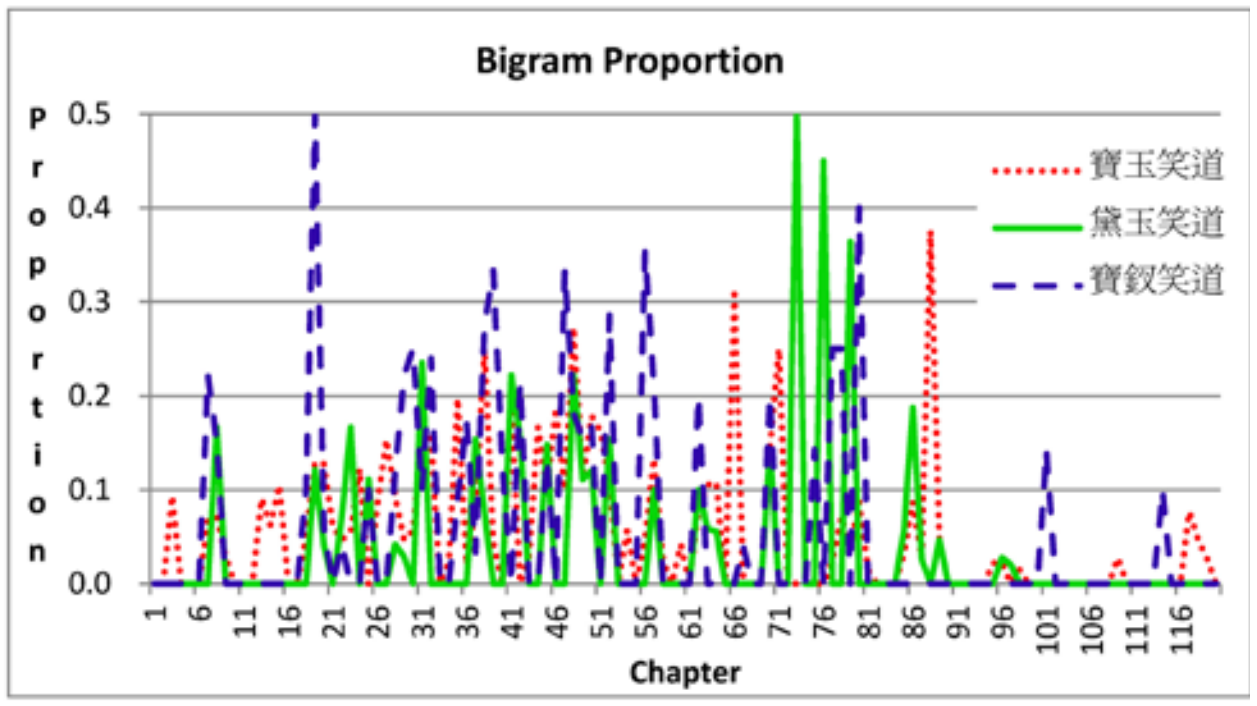

**Figure 6. Proportions of smiling of three main characters in *DRC* [11]**

Except the problem of defining “smile often”, this exploration also revealed the need to define synonyms or near synonyms for temporal analysis of keywords. There can be different ways to describe how one smiles, and “笑道” (/xiao4 dao4/) is just one of them. In English, people may smile, laugh, chuckle, giggle, etc. Likewise, there are different ways to describe smiling and laughing in Chinese. Hence, to really compare which character smile most often, we would prefer to have a more complete list of words for describing smiling.

This issue should be reminiscent of what we just described about our work with *JTTW* in the previous section, where we failed to find all monsters because we did not include all necessary keywords to search relevant statements.

## 6. Concluding Remarks and Extensions

The availability of text documents provides a key infrastructure for digital humanities, but digitizing data is just a first important step. Building software tools that help researchers extract relevant information for further investigation is a required part to make the digitized data alive and instrumental in realistic research activities.

In this paper, we present sample research projects that demonstrated the integrated applications of available data and software tools to different types of research problems in history and literature. Our examples show how software tools could help and also shed light on their current limitations.

A crucial precondition for the success of digital approach to historical studies is the reliable and complete sources of the original documents. Figure 7 portrays the appearances of some keywords in the articles of state-controlled newspapers that were published when the 228 incident occurred in 1947 in Taiwan [10]. (The leftmost digit for the “dates” on the horizontal axis represents the month, and the remaining digits are dates. Hence, 228 represents February 28th.) Lin attempted to find all reports about the incident [10], but the collection consists more of the “official” reports. Given these text files, we can extract the original statements, and will not just look at the statistics. One can investigate the government’s attitude about the incident with these reports. Nevertheless, acquiring more reports that examined this incident from different angels, e.g., those published in Japan, United States, and the communist China, will be very important for uncovering the facts.

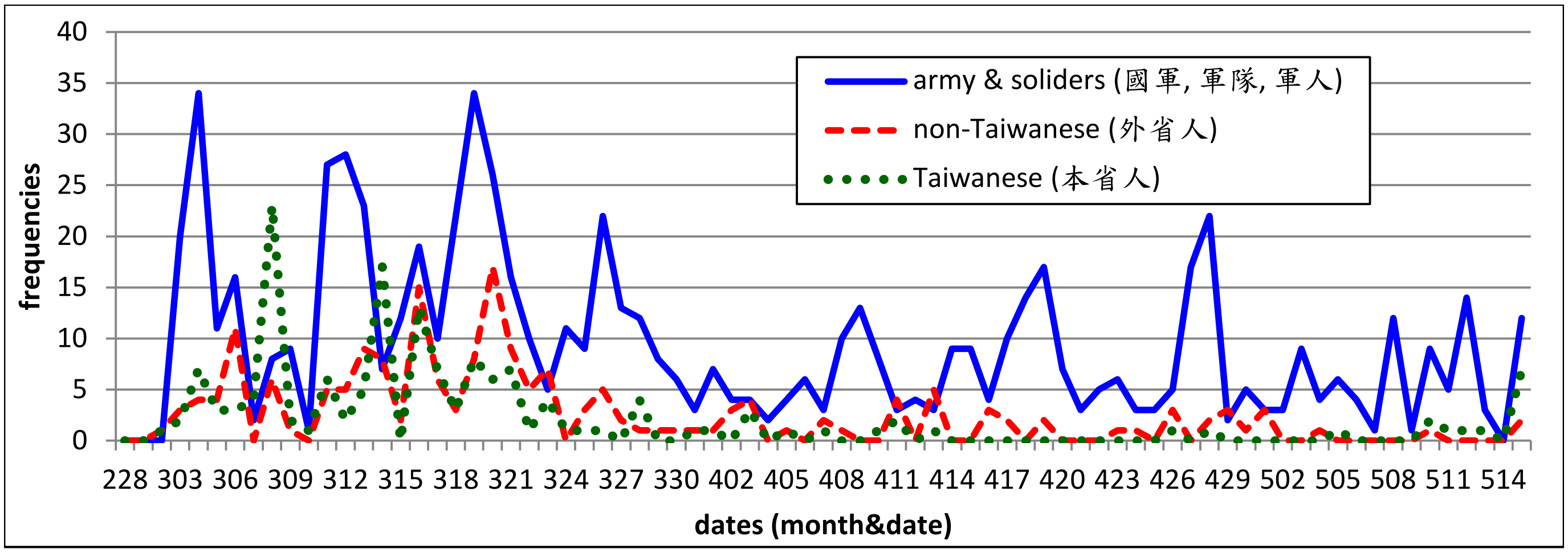

**Figure 7. Temporal analysis of some keywords in articles, in state-controlled newspapers, about the 228 incident in Taiwan**

We are expanding our programs in two fronts. We have recruited researchers for modern Chinese history and modern Chinese literature, hoping to expand the breadth of our research issues. Applying the same tools to studying issues of new areas may reveal weakness of our current approach that we did not know. We are also introducing more advanced computing technologies, such as citation identification, named entity recognition, and social network analysis, to help us do advanced spatiotemporal analysis about concepts and people.

In many of our previous projects, word frequencies played a central role in guiding historians to find relevant documents. When we counted word frequencies, some of the words may belong to certain original statements that were cited repeatedly by other authors. We should like to know the percentage of frequency that was introduced by citations so that we can analyze and interpret word frequencies more carefully.

In addition, the citations among documents may reveal the social networks among the historical figures in a study. When strengthened by opinion mining techniques, one might find parties of different opinions that involved in historical debates and their main arguments over time.

Named entities (NE) include time, locations, and person names which are important ingredients for historical studies, cf. [12]. By incorporating a temporal analysis with location information, we will open a window to investigating the propagation of concepts in terms of both time and space in ancient China. More specifically, we will empower ourselves to study when and whether a concept was propagated from the south to the north or vice versa.

NE recognition may contribute to the study of social relationships in the China Biographical Database Project at the Harvard university. Chinese words for family relati onships are much more complicated than English. With the expanding volume of the digitized data in the CBDB project, it is becoming more likely to study not only the kinship relationships but also the social networks in ancient China.

## Acknowledgments

We thank Professor Yuan-Huei Lin for the privilege to use the text files of [10]. This research was partially supported by the Ministry of Science and Technology of Taiwan under the grants 102-2420-H-004-054-MY2, 102-2420-H-004-058-MY2, 103-2918-I-004-001, 104-2221-E-004-005-MY3, and by the TOP university project of the National Chengchi University in Taiwan.

## References

[1] P. K. Bol, an unpublished report on name disambiguation, Harvard University.

[2] C.-Y. Chan and N.-X. Wang, "Isms" of the digital humanities, in *New Approaches to Historical Studies*, J. Hsiang, Ed. Taipei: National Taiwan University, 2014, pp. 219–245. (in Chinese)

[3] W.-Y. Chiu, G-T. Jin, Q.-F. Liu, and C.-L. Liu, A digital humanities research about the transition of modern China conception: The concept "Equality" as an example, in *Proceedings of the Fourth International Conference of Digial Archives and Digital Humanities*, pp. 329–373, 2012. (in Chinese)

[4] W.-Y. Chiu, G.-T. Jin, Q.-F. Liu, and C.-L. Liu. Ideas, events, and actions: The digital humanity study of the concept of "Zhuquan" formation in modern China, presented in the 20th Biennial Conference of the European Association of Chinese Studies 2014.

[5] J. R. Firth, A synopsis of linguistic theory 1930–1955, *Studies in Linguistic Analysis*, 1–32, 1957.

[6] M. K. Gold (Ed.) *Debates in the Digital Humanities*, University of Minnesota Press, 2012.

[7] J. Hsiang and F.-E. Tu, What is digital humanities, in *The Way to Digital Humanities*, J. Hsiang, Ed. Taipei: National Taiwan University, 2011, pp. 9–28. (in Chinese)

[8] G.-T. Jin, W.-Y. Chiu, and C.-L. Liu. Frequency analysis and applications of co-occurrence phrases: The origin of the "Hua-Ren" concept as an example, in *Essential Digital Humanities: Defining Patterns and Paths*, J. Hsiang, Ed. Taipei: National Taiwan University, 2012, pp. 141–170. (in Chinese)

[9] G.-T. Jin and Q.-F. Liu, *The Origin of Modern Thoughts in China* (中國現代思想的起源), Hong Kong: Chinese University of Hong Kong, 2005. pp. 346–350. (in Chinese)

[10] Y.-H. Lin (Ed.) *A Collection of the News Articles about the 228 Incident in Taiwan* (《二二八事件台灣地區新聞史料彙編》), Taipei: the 228 Memorial Foundation, 2009. (in Chinese)

[11] C.-L. Liu, G.-T. Jin, Q.-F. Liu, W.-Y. Chiu, and Y.-S. Yu, Some chances and challenges in applying language technologies to historical studies in Chinese, *International Journal of Computational Linguistics and Chinese Language Processing*, 16(1-2), pp. 27–46, 2011.

[12] W.-H. Pang, S.-P Chen, and H. Cheng. From text to data: Extracting posting data from Chinese local monographs, in *Proceedings of the Fifth International Conference on Digital Archives and Digital Humanities*, pp. 93–116, 2014. (in Chinese)

[13] W.-H. Pang, S.-G. Liu, H.-C. Tu, G.-A. Weng, and J. Hsiang, Automated name-extraction in Chinese classics: Applying PMI (Pointwise Mutual Information) segmentation to Zizhi Tongjian, in *Digital Humanities and Craft: Technological Change*, J. Hsiang, Ed. Taipei: National Taiwan University, 2014, pp. 139–163. (in Chinese)

[14] S. Schreibman, R. Siemens, and J. Unsworth (Eds.) *A Companion to Digital Humanities*, John Wiley & Sons, 2008.

[15] H. C. Tu, A text-mining approach to the authorship attribution problem of Dream of the Red Chamber, in *Digital Humanities and Craft: Technological Change*, J. Hsiang, Ed. Taipei: National Taiwan University, 2014, pp. 93–120. (in Chinese)

[16] F.-S. Wang, The possibilities and limitations of digital humanities: A historian's perspective, in *Digital Humanities and Craft: Technological Change*, J. Hsiang, Ed. Taipei: National Taiwan University, 2014, pp. 25–35. (in Chinese)

[17] Y.-C. Wang, Y.-H. Lu, R. T.-H. Tsai, Q.-F. Liu, G.-T. Jin, and C.-L. Liu, Transliteration extraction methods for the 19$^{th}$ century Chinese literature, in *Digital Humanities and Craft: Technological Change*, J. Hsiang, Ed. Taipei: National Taiwan University, 2014, pp. 121–138. (in Chinese)